\documentclass[letterpaper]{article} 
\usepackage{aaai24}  
\usepackage{times}  
\usepackage{helvet}  
\usepackage{courier}  
\usepackage[hyphens]{url}  
\usepackage{graphicx} 
\usepackage{natbib}  
\usepackage{caption} 
\usepackage{algorithm}
\usepackage{amsmath,amsfonts}
\usepackage{algorithmicx}
\usepackage{algorithm}
\usepackage{array}
\usepackage{textcomp}
\usepackage{url}
\usepackage{verbatim}
\usepackage{graphicx}
\usepackage{caption}
\usepackage{subcaption}
\usepackage{cite}
\usepackage{algpseudocode}
\usepackage{booktabs}
\usepackage{tabularx}

\makeatletter
\algnewcommand{\LineComment}[1]{\vspace{4px}\Statex \hskip\ALG@thistlm \(\triangleright\) #1\vspace{4px}}
\makeatother
\usepackage{newfloat}
\usepackage{listings}
\DeclareCaptionStyle{ruled}{labelfont=normalfont,labelsep=colon,strut=off} 
\floatstyle{ruled}
\newfloat{listing}{tb}{lst}{}
\floatname{listing}{Listing}

\title{Learning Local Constraints for Reinforcement-Learned Content Generators}

\author {
    Debosmita Bhaumik\textsuperscript{\rm 1},
    Julian Togelius\textsuperscript{\rm 2},
    Georgios N. Yannakakis\textsuperscript{\rm 1},
    Ahmed Khalifa\textsuperscript{\rm 1}
}
\affiliations {
    \textsuperscript{\rm 1}Institute of Digital Games, University of Malta, Msida, Malta\\
    \textsuperscript{\rm 2}Game Innovation Lab, New York University, New York, USA\\
    debosmita.bhaumik.22@um.edu.mt, julian@togelius.com, georgios.yannakakis@um.edu.mt, ahmed@akhalifa.com
}

\begin{document}

\maketitle

\begin{abstract}
Constraint-based game content generators that learn local constraints from existing content, such as Wave Function Collapse (WFC), can generate visually satisfying game levels but face challenges in optimizing global properties, such as playability. On the other hand, reinforcement-learning-trained generators can optimize global properties---because such properties can easily be included in reward functions---but the results can be visually dissatisfying. In this paper, we explore ways to combine these methods. Specifically, we constrain the action space of a PCGRL generator with constraints learned by WFC, effectively allowing the PCGRL generator to achieve global properties while being forced to adhere to local constraints. To better analyze how this hybrid content generation method operates, we vary the number and type of inputs, and we test whether to randomly collapse the starting state and exclude rare patterns. While the method is sensitive to hyperparameter tuning, the best of our trained generators produce visually satisfying and playable puzzle-platform game levels---such as \textit{Lode Runner} levels---with desired global properties.
\end{abstract}

\section{Introduction}

What makes a game level good? There are many factors to consider, but the most salient factors can arguably be divided into functional aspects and aspects of visual aesthetics. Functional aspects relate to what the player can do in the level, e.g., can they finish it, which skills are needed, and which items can be reached. Visual aesthetics are in themselves multifaceted, but typically, a game has a specific visual style, and levels that do not adhere to this style look broken. While visual aesthetics relate to both global and local aspects of a level, functional aspects are generally global. In particular, whether a level can be finished or an item reached can only be evaluated in the context of the whole level.

Self-supervised learning approaches to image generation are generally very good at capturing visual aesthetics, and this capability extends to level generation if enough data is available~\cite{volz2018evolving,snodgrass2016learning,summerville2016super}. However, they do not inherently capture the functional aspects of a game level, perhaps because this is not part of their learning signal. Reinforcement learning approaches, on the other hand, can be used to learn level generation models that capture functionality well if the reward is made explicitly dependent on such functionality~\cite{khalifa2020pcgrl,shu2021exppcgrl,earle2021control}. Unfortunately, this often comes at the expense of visual aesthetics, as levels generated via reinforcement learning can be downright ugly (Figure~\ref{fig:pcgrl_levels}).

The question naturally arises whether we can combine self-supervised and reinforcement learning methods to learn level generators that generate functional levels that adhere to specific visual styles. This may or may not take the form of learning local patterns via self-supervised learning and global structure via reinforcement learning. This paper proposes one specific method for doing exactly this, combining the Wave Function Collapse (WFC)~\cite{gumin2016wfc} algorithm for learning local patterns and reinforcement learning for learning to produce playable levels. The specific way these methods are combined is by letting WFC limit what action the RL model can take. 

For game level generation, a generated level must be playable. Often, levels generated using PCG via machine learning (PCGML)~\cite{pcgml} look similar to the training human-made levels, but do not guarantee functionality. The obvious reason is that functionality does not depend on or is not related to the visual similarity or aesthetics. For example, a Super Mario Bros level with some randomly scattered floor tiles in the sky may look messy, but the level is still functionally complete if there exists a path to the goal. The other way also holds; a level that looks like it was designed by humans can be non-functional if the path does not exist. We are exploring how to generate playable levels that carry visual similarities with the given input using a reinforcement learning (RL) approach. RL methods have shown great success in generating content, but incorporating the visual similarity measure in a reward function is not very straightforward. 

In this work, we are combining WFC with an RL-based approach similar to the work of \citet{babin2021leverging}; however, we apply a PPO-based method instead of an evolution-based approach to train the agent. We use a PPO-based PCGRL framework and constrain the action space of PCGRL using the local rules derived by WFC. We study the effects of the algorithm's hyperparameters on the final generated content. We investigate the size of input data, the diversity of input data, excluding less frequent patterns from input data, and the effect of the starting state. Finally, we ran a small user study to compare our method to using WFC or an RL approach alone to understand how humans perceive the generated content.

\section{Background}

\subsection{Procedural Content Generation via Machine Learning (PCGML)}\label{pcgml}
PCGML~\cite{pcgbook} research focuses on the generation of game content (such as maps, quests, levels, music, narrative, etc) using input examples. In this approach, a machine learning model is trained using the input data. The model tries to learn the underlying distribution of the training data; afterwards, the trained model is used to generate new content. Various machine learning approaches have been explored for automated content generation, which range from Markov models~\cite{snodgrass2016learning}, LSTM networks~\cite{summerville2016super}, Generative Adversarial Networks (GANs)~\cite{volz2018evolving,schrum2020interactive}, AutoEncoders~\cite{sarkar2020controllable}, to recent Large Language Models (LLMs)~\cite{Todd2023GenerationLLMs, Nasir2023Practicalpcg}. 

\subsection{Procedural Content Generation via Reinforcement Learning (PCGRL)}\label{pcgrl}
PCGRL~\cite{khalifa2020pcgrl} treats the content generation problem as a Markov Decision Process (MDP), where the agent is trained to select an action that leads towards a goal; in return, it receives a reward indicating how good or bad the action is. This constitutes an iterative approach to the content generation problem, where the content is generated in a step-by-step manner rather than a one-shot process. One advantage of RL-based methods compared to other machine learning methods is that RL-based methods do not require training data. Instead, a reward function is used to guide the generation process, which can help the trained agent to learn more complex concepts such as playability. 

\citet{khalifa2020pcgrl} introduced an RL-based framework for 2D game levels, where starting with a random level, the RL agent iteratively modifies the level towards a certain goal. \citet{earle2021control} proposed a controllable RL-based generator, where they used control parameters for training the agent. At inference time, using the control parameter, users can generate a variety of content from a single generator. \citet{jiang2022control3d} applied an RL-based controllable generator on a more complex 3D game environment. \citet{Gisslen2021adversarial} applied an adversarial RL approach for PCG. They adversarially trained a PCGRL generator using an RL-based solving agent for generating novel game environments. In some other direction, \citet{shu2021exppcgrl} combined PCGRL with experience-driven PCG to generate personalized game content.


\subsection{Wave Function Collapse (WFC)}\label{wfc}
WFC was initially proposed by~\citet{gumin2016wfc} for generating images and tile maps that have pattern similarities with a given input. The algorithm takes pixel or tile-based example input, divides the input into NxN patterns, and extracts the local relations between these patterns, which define the constraints for the algorithm. Following these constraints, the algorithm produces outputs having pattern similarities with the given input. Since its inception, it has become popular among game designers as well as game researchers due to the aesthetically pleasing output and the need for a small amount of input data. WFC has been applied and adapted in various games such as Bad North~\cite{stalberg2018badnorth}, Townscaper~\cite{tommy2022townscaper}, Caves of Qud~\cite{bucklew2022caves}, etc. 

Several academic studies have explored WFC in different ways. \citet{karth2017wfcwild} investigate the use of WFC as a constraint-solving PCG approach. In a follow-up work, they explore different ways to extend the algorithm and overcome its limitations, such as using VQ-VAE as a tile representation~\cite{karth2019wfcdis} and using positive and negative examples as inputs~\cite{karth2022wfcml}. \citet{sandhu2019enhance} explore the idea of integrating design constraints as a general framing of WFC constraints and investigate their effectiveness. Instead of using a grid structure, a graph structure can be used to expand on the functionality of the method and reduce its limitations~\cite{cooper2023sturgeon,kim2019graph}. In another study, \citet{moller2020growing} applied WFC on a growing grid rather than a fixed-sized grid to overcome the limitation of having a specific level size. \citet{lee2020precomputing} used a constraint-based approach by pre-computing player paths and transition states; then built WFC-inspired tile neighboring constraints that ensure the reachability of the generated levels. To improve the playability of WFC, \citet{facey2024spacetime} used a 3-dimensional WFC, where the third dimension, called space-time, contains the steps to solve the level.
\citet{babin2021leverging} combined a reinforcement learning approach with WFC for generating playable Super Mario levels to overcome functionality problems. They applied an ES-based optimization approach to train an RL agent that replaces the tile and action selection of WFC.

\subsection{Lode Runner}
Lode Runner is a platformer-puzzle game, published by Broderbund in 1983. The game is about collecting gold without getting killed by the enemies. The player can walk on platforms, through ropes, and climb ladders, but they cannot jump. Besides traversing a level, the player can dig holes in bricks to make a new path or use the holes to trap/kill enemies. Though Lode Runner is not a popular choice for game AI research, its spatial relations between different tiles and puzzle-like nature make it a good candidate for our experiment, as it is easy to detect a level that does not follow Lode Runner's structure, and it has hard connectivity and functionality constraints that need to be achieved.

\citet{snodgrass2016learning} trained a multi-dimensional Markov model to produce levels for Super Mario Bros, Lode Runner, and Kid Icarus. \citet{steckel2021illuminating} used a GAN with the MAP-Elites algorithm to generate diverse playable levels of Lode Runner. \citet{sorochan2021lsmt} trained an LSTM on the player path of Lode Runner levels and used it as a level generator. \citet{snodgrass2020multi} combined variational autoencoders with example-driven binary space partitioning to blend and generate levels from multiple domains, including Lode Runner. \citet{thakkar2019autoencoder} applied evolution on the latent space of autoencoders and variational autoencoders to generate Lode Runner levels. 

\section{Wave Function Collapse via Reinforcement Learning (WFCRL)}

\begin{figure}[!tb]
    \centering
    \includegraphics[width=0.5\linewidth]{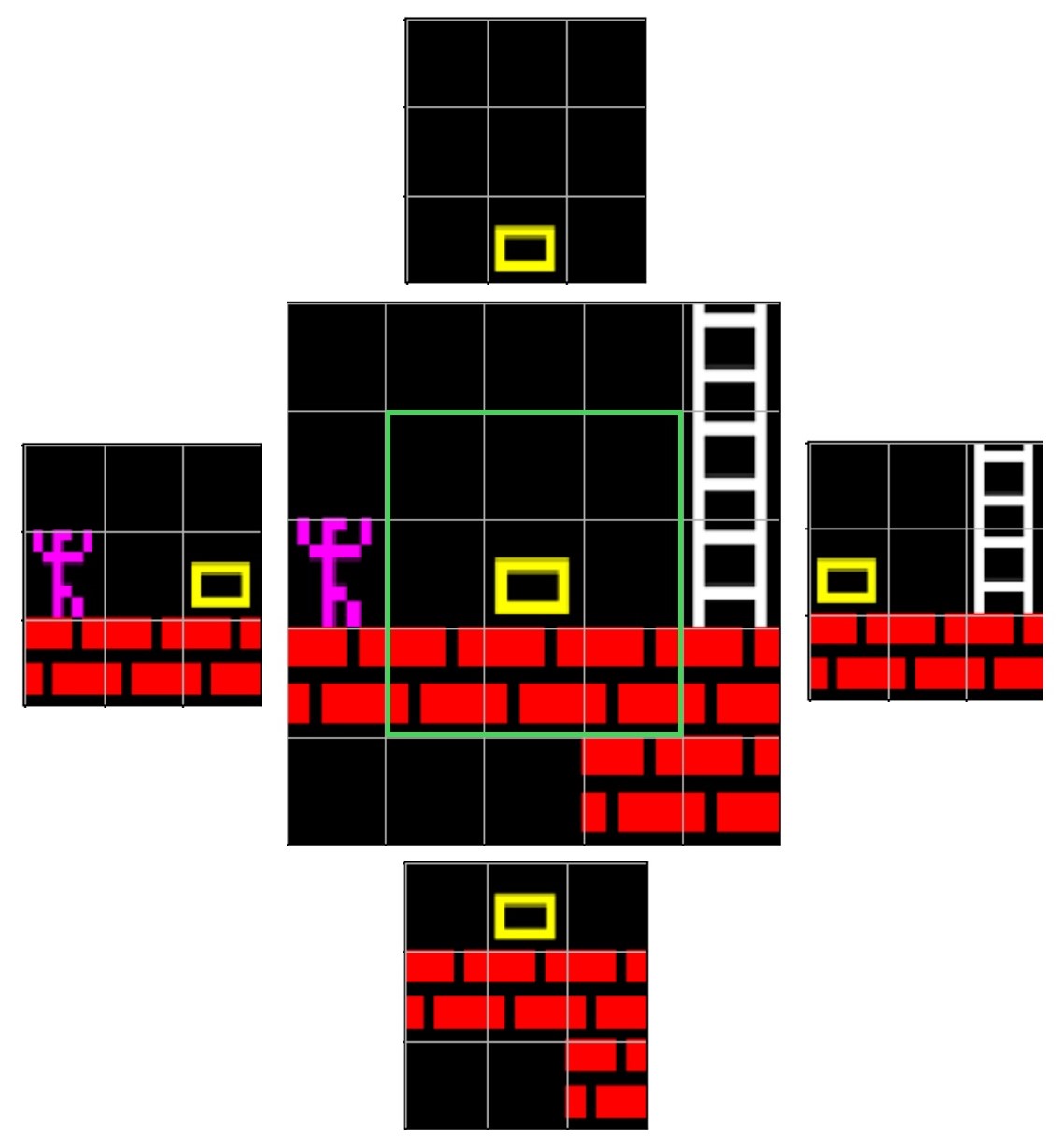}
    \caption{Adjacency relation of the marked pattern using a $3\times3$ window. The selected pattern is marked by a green border, and the $3\times3$ neighbor patterns in four cardinal directions are shown on the side.}
    \label{fig:adjacency}
\end{figure}

\begin{algorithm}[t]
\caption{Pseudo Code of a full level generation using the WFCRL framework. The observation ($lvl$ and $loc$), action ($pattern$), and reward ($reward$) get added to the training trajectories ($train\_trajectory$), which is used to train the RL agent.}\label{alg:WFCRL}
\begin{algorithmic}[1]
\LineComment{Initialize empty level with all possible patterns}
\State $patterns \gets extract\_patterns(input)$\label{algo:patterns}
\State $adj\_rules  \gets find\_adjacency\_rules(patterns)$\label{algo:adjacency}
\State $lvl  \gets empty\_grid(patterns)$\label{algo:empty}
\LineComment{Assign a single player pattern to the level}
\State $loc \gets random(lvl)$\label{algo:rand_loc}
\State $player\_patterns  \gets get\_player\_patterns(patterns)$\label{algo:rand_player_1}
\State $pattern \gets random(player\_patterns)$\label{algo:rand_player_2}
\State $lvl \gets apply\_pattern(loc, pattern, lvl, adj\_rules)$\label{algo:apply_player}
\State $remove\_patterns(lvl, player\_patterns )$\label{algo:clear_player}
\LineComment{Collapse the level using WFCRL Framework and collect the training trajectory}
\State $train\_trajectory \gets [\ ]$
\While{not $cell\_collapsed(lvl)$}\label{algo:loop}
\State $loc \gets  next\_cell\_to\_collapse(lvl)$\label{algo:next}
\State $av\_patterns \gets get\_valid\_patterns(loc, lvl)$\label{algo:av_patterns}
\If{$len(av\_patterns)$ is $0$}\label{algo:contra_1}
\State \Return contradictions error\label{algo:contra_2}
\EndIf\label{algo:contra_3}
\State $pattern  \gets RL\_agent(loc, lvl, av\_patterns)$\label{algo:rl}
\State $n\_lvl \gets apply\_pattern($ \\
    \hspace*{2em} $loc, pattern, lvl, adj\_rules)$\label{algo:resolve}
\State $reward  \gets RL\_reward(n\_lvl, lvl)$\label{algo:reward}
\State $add(train\_trajectory, $ \\
    \hspace*{2em} $[lvl, loc, pattern, reward])$
\State $lvl \gets n\_lvl$
\EndWhile
\LineComment{Return the fully collapsed level and the training trajectory}
\State \Return $lvl, train\_trajectory$\label{algo:lvl}
\end{algorithmic}
\end{algorithm}

In this paper, we integrate the constraint-solving power of WFC with the PCGRL framework to generate visually pleasing and functional levels for the platformer game Lode Runner. Similar to the work by~\citet{babin2021leverging}, we combine RL with WFC for generating playable levels where the RL selects the value of the next tile. One key difference is that \citeauthor{babin2021leverging} used an ES-based optimization approach to train the RL agent, while we are using a PPO-based RL agent. In gradient-based approaches, incremental rewards work better, so we use a step-wise playability reward compared to the end playability percentage. Another difference is the domain itself; \citeauthor{babin2021leverging} focused on generating linear levels for Super Mario Bros, while we are focused on the Puzzle game Lode Runner. Lode Runner is a harder problem to solve compared to Mario, where playability is easier to achieve~\cite{snodgrass2016learning}. We also extend their work by investigating some of the factors that affect the quality and the visual aesthetics of the generated content. For example, the size of the input data, the diversity of the input data, the presence and exclusion of less frequent patterns, and finally, training from an empty state vs a random collapsed starting state. 

WFC operates on the tiles by treating them as pixels. The framework takes input level(s) and extracts $N \times N$ unique tile patterns (where N is the size of local constraints; usually 2 or 3 works best) present in the input level(s). These unique patterns and their neighboring patterns in four cardinal directions define the adjacency rules. Figure~\ref{fig:adjacency} displays $3 \times 3$ adjacency relations for a selected pattern. These adjacency relations tell what patterns can be placed as neighbors of a selected pattern in four cardinal directions. 
The pattern dataset acts as the action space for the RL agent. At each step, using adjacency rules, WFC generates a list of valid patterns for the selected cell, which is used for action masking.

Algorithm~\ref{alg:WFCRL} shows the main steps for running the algorithm from the start state till a level is generated. At initialization, WFC extracts the patterns and adjacency rules from the input level(s) (lines \ref{algo:patterns} and \ref{algo:adjacency}). We create an empty grid-like level, where each cell contains the possible patterns that can be placed at that location (line \ref{algo:empty}). Initially, all patterns are available to be placed in any location. WFC picks the most constrained cell (i.e., the cell with the least number of available patterns (line \ref{algo:next})) and provides it, the current state of the level, and the available patterns to the RL agent, which selects one of the available patterns (line \ref{algo:rl}). After the agent selects the pattern, WFC applies the pattern and propagates that selection to the whole level by removing any patterns that will conflict with the adjacency relation (line \ref{algo:resolve}). If at any point the most constrained cell does not have any more choices, WFC raises a contradiction error, which indicates failure to generate the level (lines \ref{algo:contra_1}, \ref{algo:contra_2}, and \ref{algo:contra_3}). If the propagation is completed successfully, the framework calculates a reward signal that signifies how close the new level is to playability from the previous level (line \ref{algo:reward}). This process continues until all the cells are collapsed or a contradiction occurs during the propagation (line \ref{algo:loop}). Before the framework starts, we place a random player pattern (lines \ref{algo:rand_player_1} and \ref{algo:rand_player_2}) at a random location (line \ref{algo:rand_loc}) and propagate it through the level (line \ref{algo:apply_player}), then remove all the patterns that could add an additional player (line \ref{algo:clear_player}).

\begin{figure}
    \centering
    \begin{subfigure}[b]{0.95\linewidth}
        \centering
        \includegraphics[width=\linewidth]{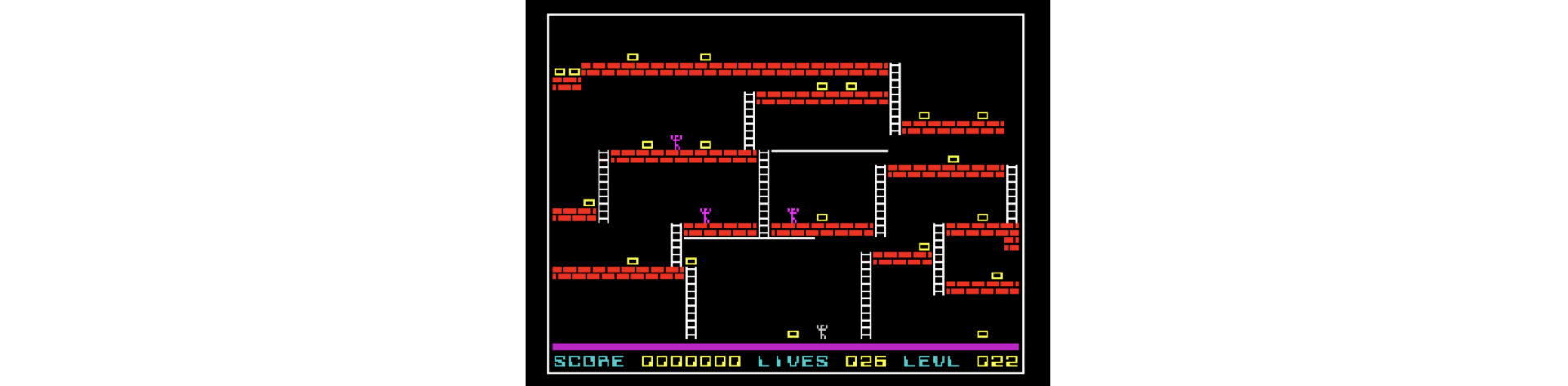}
        \caption{The input level for the Single Level experiment.}
        \label{fig:si_inputs}
    \end{subfigure}
    \begin{subfigure}[b]{0.95\linewidth}
        \centering
        \includegraphics[width=\linewidth]{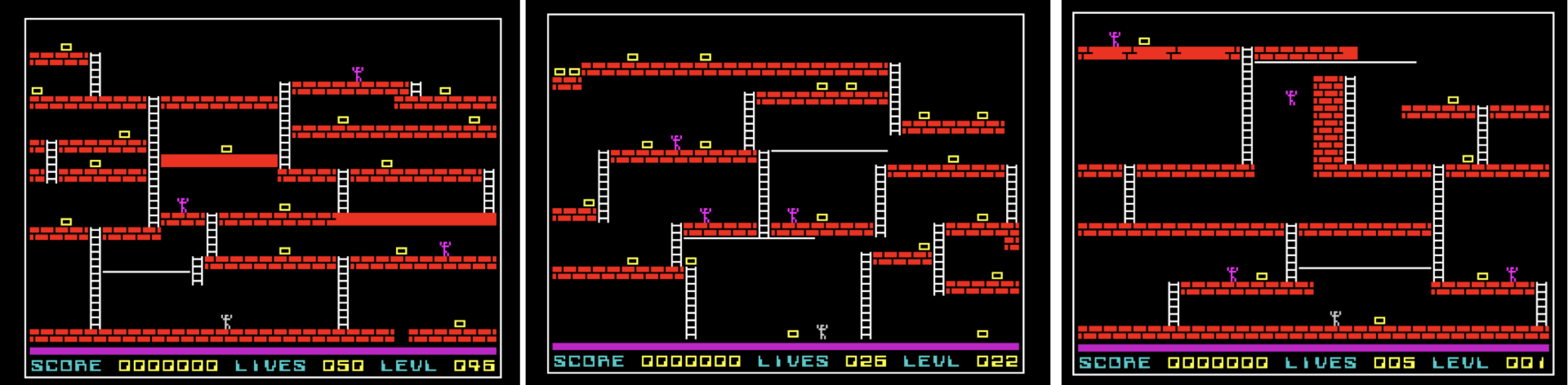}
        \caption{The input levels for the low TP-KLDiv experiment.}
        \label{fig:mi_inputs}
    \end{subfigure}
    \begin{subfigure}[b]{0.95\linewidth}
        \centering
        \includegraphics[width=\linewidth]{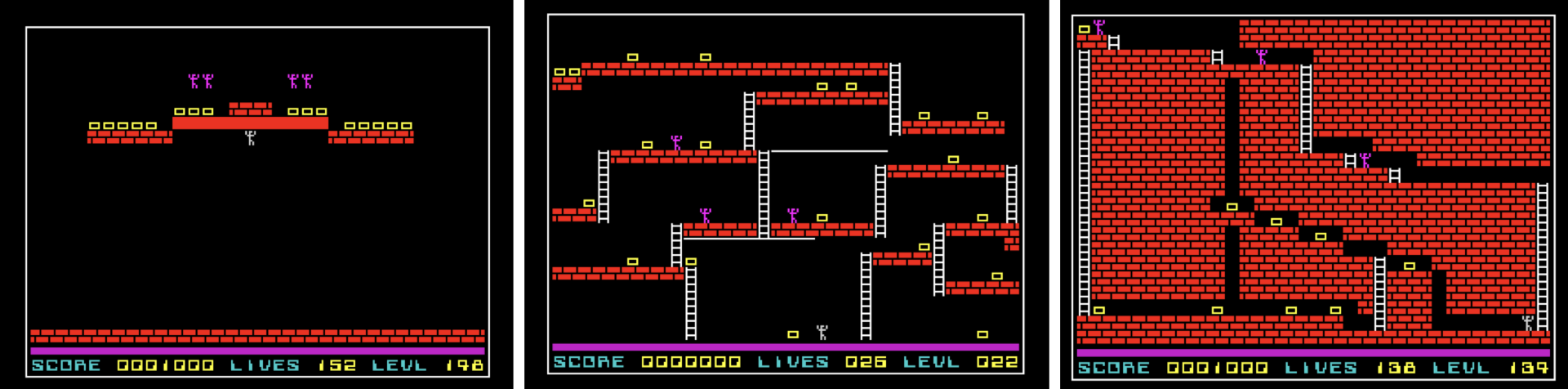}
        \caption{The input levels for the high TP-KLDiv experiment.}
        \label{fig:div_mi_inputs}
    \end{subfigure}
    \caption{The different input levels for all the experiments where Level 22 is used in all of them and other levels are added based on the diversity of the levels, calculated using the Tile Pattern KL-Divergence (TP-KLDiv) score.}
    \label{fig:input_levels}
\end{figure}

\begin{figure*}
    \centering
    \begin{subfigure}[b]{0.49\linewidth}
        \centering
        \includegraphics[width=\linewidth]{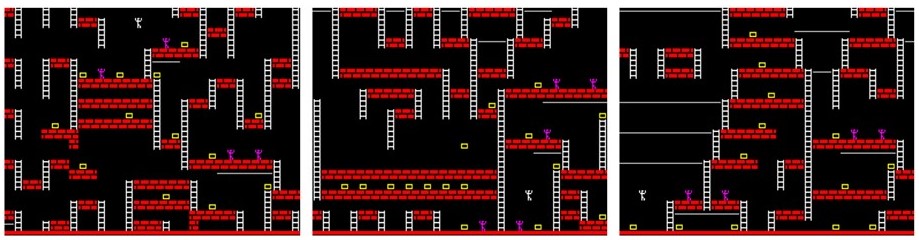}
        \caption{No Restriction}
        \label{fig:si_no_res}
    \end{subfigure}
    \begin{subfigure}[b]{0.49\linewidth}
        \centering
        \includegraphics[width=\linewidth]{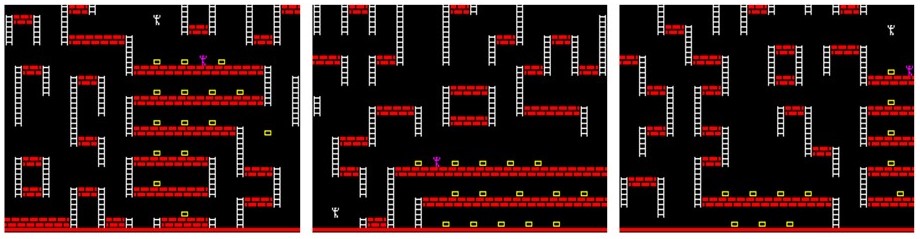}
        \caption{Excluding Rare Patterns}
        \label{fig:si_rr}
    \end{subfigure}
    \begin{subfigure}[b]{0.49\linewidth}
        \centering
        \includegraphics[width=\linewidth]{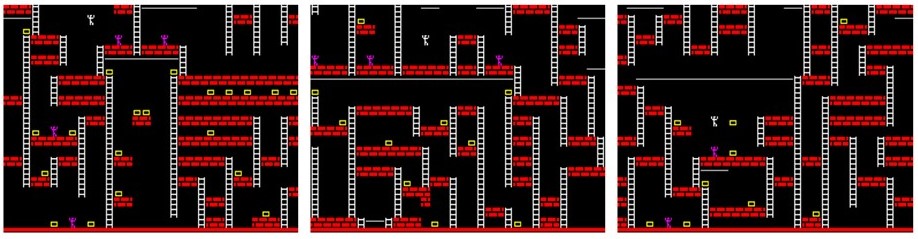}
        \caption{Random Collapsed Starting State}
        \label{fig:si_rc}
    \end{subfigure}
    \begin{subfigure}[b]{0.49\linewidth}
        \centering
        \includegraphics[width=\linewidth]{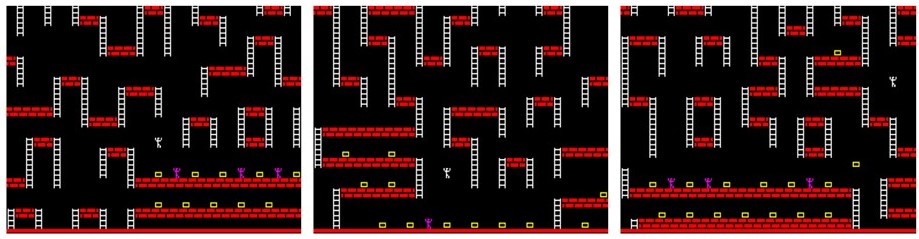}
        \caption{Excluding Rare Patterns + Random Collapsed Starting State}
        \label{fig:si_rr_rc}
    \end{subfigure}
    \caption{Levels generated using single input frames}
    \label{fig:single_input}
\end{figure*}

\begin{figure*}
    \centering
    \begin{subfigure}[b]{0.49\linewidth}
        \centering
        \includegraphics[width=\linewidth]{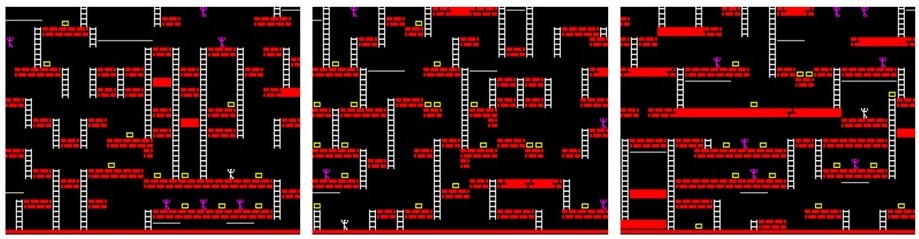}
        \caption{No Restriction}
        \label{fig:mi_no_res}
    \end{subfigure}
    \begin{subfigure}[b]{0.49\linewidth}
        \centering
        \includegraphics[width=\linewidth]{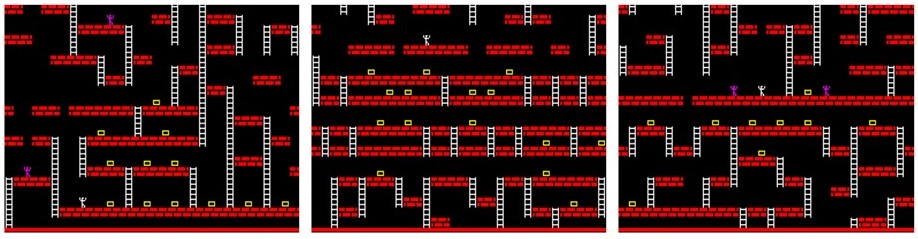}
        \caption{Excluding Rare Patterns}
        \label{fig:mi_rr}
    \end{subfigure}
    \begin{subfigure}[b]{0.49\linewidth}
        \centering
        \includegraphics[width=\linewidth]{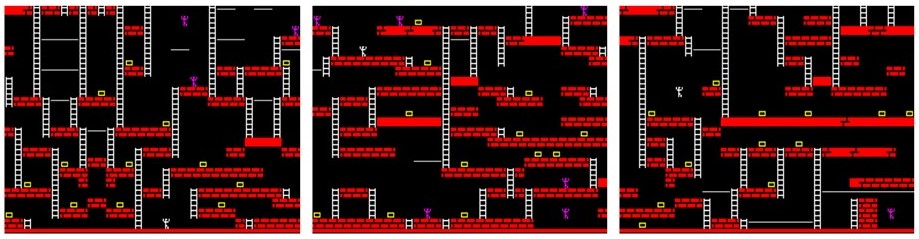}
        \caption{Random Collapsed Starting State}
        \label{fig:mi_rc}
    \end{subfigure}
    \begin{subfigure}[b]{0.49\linewidth}
        \centering
        \includegraphics[width=\linewidth]{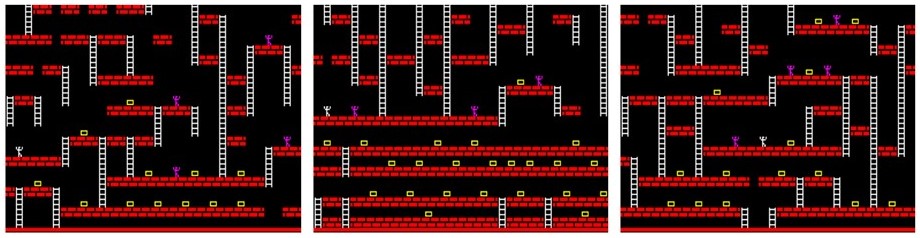}
        \caption{Excluding Rare Patterns + Random Collapsed Starting State}
        \label{fig:mi_rr_rc}
    \end{subfigure}
    \caption{Levels generated using multiple input frames.}
    \label{fig:multiple_input}
\end{figure*}

\begin{figure*}
    \centering
    \begin{subfigure}[b]{0.49\linewidth}
        \centering
        \includegraphics[width=\linewidth]{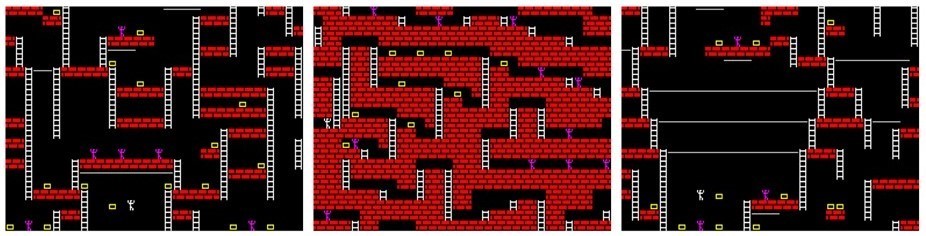}
        \caption{No Restriction}
        \label{fig:div_mi_no_res}
    \end{subfigure}
    \begin{subfigure}[b]{0.49\linewidth}
        \centering
        \includegraphics[width=\linewidth]{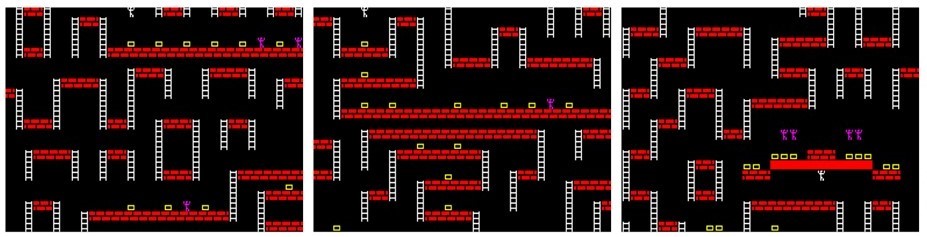}
        \caption{Excluding Rare Patterns}
        \label{fig:div_mi_rr}
    \end{subfigure}
    \begin{subfigure}[b]{0.49\linewidth}
        \centering
        \includegraphics[width=\linewidth]{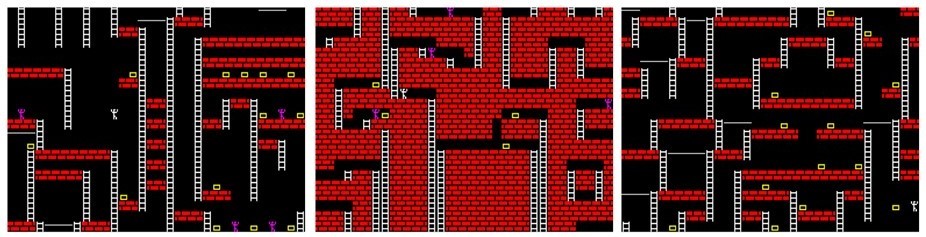}
        \caption{Random Collapsed Starting State}
        \label{fig:div_mi_rc}
    \end{subfigure}
    \begin{subfigure}[b]{0.49\linewidth}
        \centering
        \includegraphics[width=\linewidth]{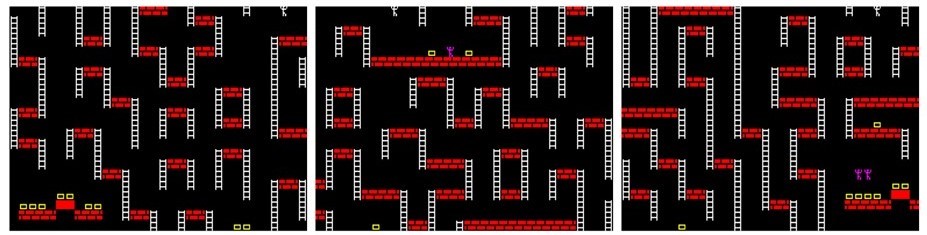}
        \caption{Excluding Rare Patterns + Random Collapsed Starting State}
        \label{fig:div_mi_rr_rc}
    \end{subfigure}
    \caption{Levels generated using highly diverse multiple input frames.}
    \label{fig:div_multiple_input}
\end{figure*}

\subsection{WFC Rule Learning \& Constraints}

As discussed above, we use the full loop of the WFC, but we replace the pattern selection strategy from the original WFC with the RL agent. Instead of using the prior distribution of the patterns, we allow the RL agent to select the pattern that will lead towards functionality. WFC learns the adjacency constraints (lines \ref{algo:patterns} and \ref{algo:adjacency}) and restricts the RL agent's action space (lines \ref{algo:av_patterns} and \ref{algo:rl}), such that the generated levels will have a similar look to the input data. When the RL agent selects a pattern, WFC will propagate that selection all over the level and remove any patterns that will not work with that selection (line \ref{algo:resolve}). This will prevent the RL agent from selecting any invalid actions in the next step.

\subsection{RL Observation}

We are following a representation that is similar to the narrow representation from the PCGRL framework~\cite{khalifa2020pcgrl}. At each step, a location and the current level are given to the agent (line \ref{algo:rl}). The level is a 3D array of size $l \times w \times t$ where $l$ and $w$ are the dimensions of the level and $t$ is the total number of different tiles available. If a cell is collapsed, the number of available tiles in that cell will be 1. The current location is provided as a (row, column) index by the WFC. We follow the same idea as the narrow representation and translate the input observation such that the location becomes the center of the observation. This makes the transformed level twice the size of the actual level. As player placement is not handled by the RL agent, the player channel is also not included in the observation space. The player location is treated as an empty tile in the observation. This leads the framework to learn to have a connected level rather than where the player starts from.

\subsection{RL Action Space}

The action space is defined by the patterns obtained from the WFC (line \ref{algo:patterns}); the number of actions ranges from 0 to $n-1$, where $n$ is the total number of patterns in the input level. Since not all the patterns are available for the agent, we mask the output actions so that only the available actions can be selected from (line \ref{algo:rl}). Similarly to the work of \citet{huang2022actionmask}---where they have shown the effectiveness of invalid action masking over the non-masked actions for Super Mario Bros level generation---we are masking the actions coming out from the RL agent. At each step, WFC generates a list of available patterns for the location to be modified (line \ref{algo:av_patterns}). The available patterns information is converted into masked actions and sent to the agent. The masked actions are an array of size $n$, matching the number of patterns; for available patterns, the value is set to 1, and 0 otherwise. Using masking not only prevents the agent from selecting invalid actions but also prevents gradients from updating the network based on the invalid action output.

\subsection{RL Reward}

The RL loop tries to optimize the playability of the generated level; that is, all gold should be reachable from the player's location. The playability is measured as the ratio of the number of reachable gold pieces to the total number of gold pieces in the level. An automated game-playing agent, which follows a simplified version of the game mechanics, is used to evaluate the playability. The agent tries to find the number of reachable gold pieces from the player's location using a Breadth-First Search algorithm following the game mechanics. To keep the measure simple and quick, we did not include the digging ability or automate enemy movements. Our method also does not take into account the order of collecting gold and does not handle deadlocks; it is more of a perceived playability rather than the actual playability of Lode Runner. $22.67\%$ of original Lode Runner levels are playable using our measure. In this paper, by playability we refer to the perceived playability of the levels.

The RL reward function incentivizes generating at least $4$ gold pieces and penalizes the decrease in reachable gold pieces, encouraging placing gold reachable from the player’s location. If the selected action improves the playability by making the golds reachable from the player's position, a positive reward is given proportional to the number of new reachable gold. Decreasing the reachability leads to a negative reward (proportional to the decrease in the number of new reachable gold), indicating a bad action (line \ref{algo:reward}). If the selected action results in any contradiction (line \ref{algo:contra_2}) during the propagation process of WFC, a big negative reward ($-20$) is given (we do not use any backtracking). This helps the agent to learn to take actions that will not lead to contradiction. In this framework, we have two termination conditions: the level is fully collapsed (line \ref{algo:lvl}), or there is a contradiction in the propagation process of WFC (line \ref{algo:contra_2}).

\section{Experiments}

We test all the different input parameters of the framework to understand their effect on the final output. We focus on three main parameters: the input levels, the learned patterns, and finally the starting state. In the following subsections, we will discuss the different experiments related to them.

\subsection{Input levels}

In this framework, the visual aesthetic of the generated level is dependent on the input level(s). To explore how the input level influences the generated level, we have used a single input level as well as multiple input levels. For our experiments, we used Lode Runner levels from the Video Games Level Corpus (VGLC)~\cite{summerville2016vglc}. Additionally, for multiple input levels, the diversity of the input levels also affects the output levels. Therefore, we picked two different sets, one set having minimum diversity (i.e., containing levels that look similar to each other) and the other having high diversity (i.e., containing levels that look very different from each other). Diversity is calculated using the average tile-pattern KL-divergence (TP-KLDiv) score~\cite{lucas2019tile} between all levels in pairs. The selected levels can be seen in Figure~\ref{fig:input_levels}. To extend WFC to work with multiple inputs, WFC extracts $N\times N$ unique tile patterns and the adjacency relations of each input separately. These patterns and the adjacency rules are then combined to create the final tile-pattern dataset and adjacency constraints.

\subsection{Learned Patterns}

While constructing the pattern dataset from input level(s), we found that some patterns have higher occurrences, whereas some patterns have very low occurrences in the input level. The patterns that appear only once in the input level (i.e., named as \textit{rare}) are low-frequency patterns and have very strict adjacency rules. Such strict rules usually push the level to collapse to a smaller set of options, which leaves the RL agent with fewer options to make a level playable or avoid contradictions. Longer training can help the agent to avoid such patterns, but often a selected pattern does not lead to a contradiction immediately, making it harder to connect the events and leading to a complex reward assignment problem. The inclusion (or exclusion) of such rare patterns is one hyperparameter that we consider for the experiment. To exclude rare patterns, we discard all patterns that have a single occurrence in the dataset. But in single-input experiments, removing rare patterns with player tiles failed to propagate the player placement, as those are the only patterns that can be placed at the neighboring cells of the player tile. To handle this, for single-input experiments without rare patterns, the patterns having player tiles are not excluded from the pattern dataset.

\begin{figure}
    \centering
    \includegraphics[width=0.95\linewidth]{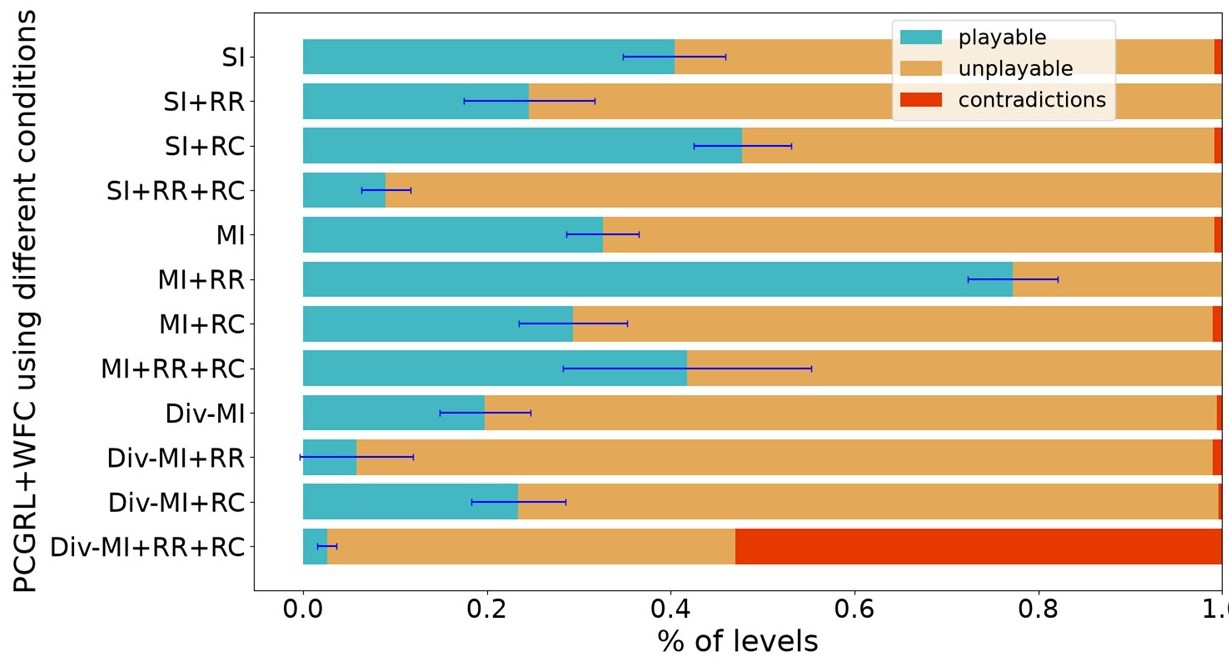}
    \caption{Compares the playability, unplayability and contradiction of levels generated using different conditions: single input (SI), multiple inputs (MI), highly diverse multiple input (Div-MI), excluding rare patterns (RR), and random collapsed starting state for training (RC).}
    \label{fig:playable}
\end{figure}

\subsection{Starting State}
Our generation process starts with an empty level; sometimes this makes the agent learn a narrow range of levels that are similar to each other. To allow the RL to explore different states and learn more generalizable policies, we start from a random, partially collapsed state instead of an empty one. Some parts of the level are already set using normal WFC, and the agent has to continue building the level on top of it. This random collapse is only done during the training phase; it is not applied during inference.

\subsection{Setup}

Our framework is implemented as an  OpenAI Gym interface~\cite{brockman2016gym}. For our experiments, we use Lode Runner level generation as the problem, where the goal is to generate playable Lode Runner levels of size $32 \times 22$. For the input levels, we use Lode Runner levels from the Video Game Level Corpus (VGLC)~\cite{summerville2016vglc}, with a $3 \times 3$ window as the pattern size. We opted for a $ 3 \times 3$ size as it can manage to capture the spatial dependencies well in Lode Runner. A smaller size cannot capture the dependencies well, while a bigger size increases unique patterns, resulting in strict adjacency rules that cause frequent contradictions. For training, we used Maskable Proximal Policy Optimization (Maskable PPO), a variation of Proximal Policy Optimization (PPO) from Stable-Baselines3-contrib~\cite{stablebaselines3}. Our policy uses the same body for both the action and value heads, and it is made of $3$ convolution layers followed by $2$ fully-connected layers. We ran an ablation study varying the above-mentioned factors, which made a total of $12$ experiments. For random-starting state experiments, we initially collapsed $10$ locations (roughly $1.5\%$ of cells) and propagated using normal WFC. Each of our experiments runs for $5$ million timestamps. We trained $5$ different models for each experiment to show the stability of the training process. 

We denote the different experiment settings as single input by `SI', multiple input by `MI', highly diverse multiple input as `div-MI', excluding rare patterns as `RR', and random starting state as `RC'. For the single input, we used a traditional-looking Lode Runner level for that, while for multiple input, we selected two additional levels such that the subset has the smallest TP-KLDiv or the highest TP-KLDiv, which can be seen in Figure~\ref{fig:input_levels}.

\section{Results}

We generate $100$ levels using each trained model and compare the playability and diversity of the generated levels from different models. Figures~\ref{fig:single_input}, ~\ref{fig:multiple_input}, and~\ref{fig:div_multiple_input} display playable levels generated using single, multiple, and diverse-multiple inputs, respectively. We can notice that the generated levels follow a similar structure to the input levels, with high similarity between all the generated levels except for the models trained on diverse inputs. Models trained using diverse inputs preferred to stick to a specific style and continue the generation. For example, certain levels have a high amount of solid tiles, or others have long ropes. The trained high-diversity model failed to combine these different styles together; we believe that might be due to the adjacency constraints from one level usually not leading to another level.

\begin{figure}
    \centering
    \includegraphics[width=0.95\linewidth]{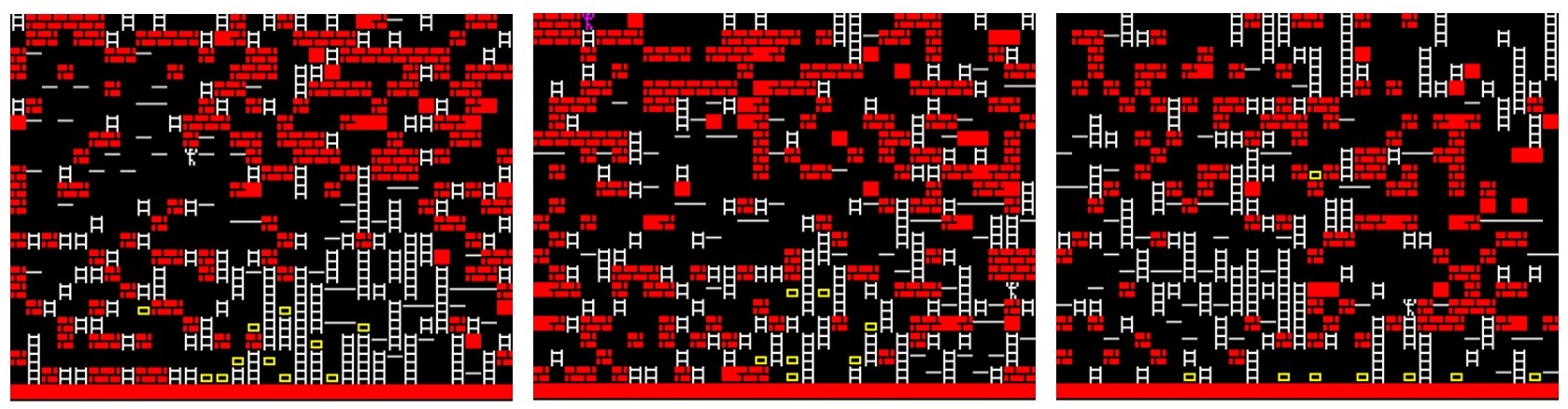}
    \caption{Example of the levels generated by the PCGRL agent alone.}
    \label{fig:pcgrl_levels}
\end{figure}

\begin{figure*}
     \centering
     \begin{subfigure}[b]{0.32\textwidth}
         \centering
         \includegraphics[width=\linewidth]{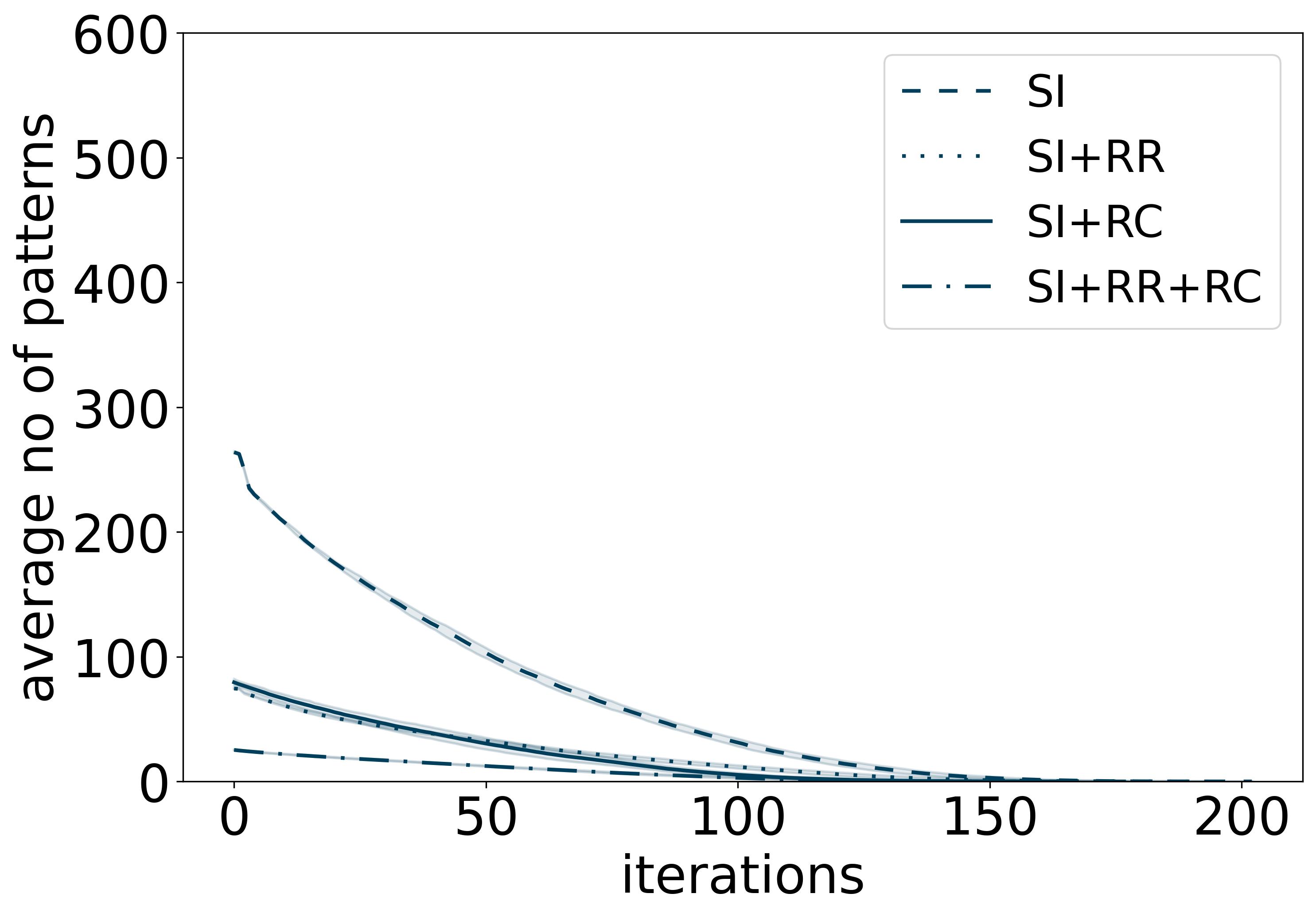}
         \caption{Single Input}
         \label{fig:no_patterns_single_input}
     \end{subfigure}
     \begin{subfigure}[b]{0.32\textwidth}
         \centering
         \includegraphics[width=\linewidth]{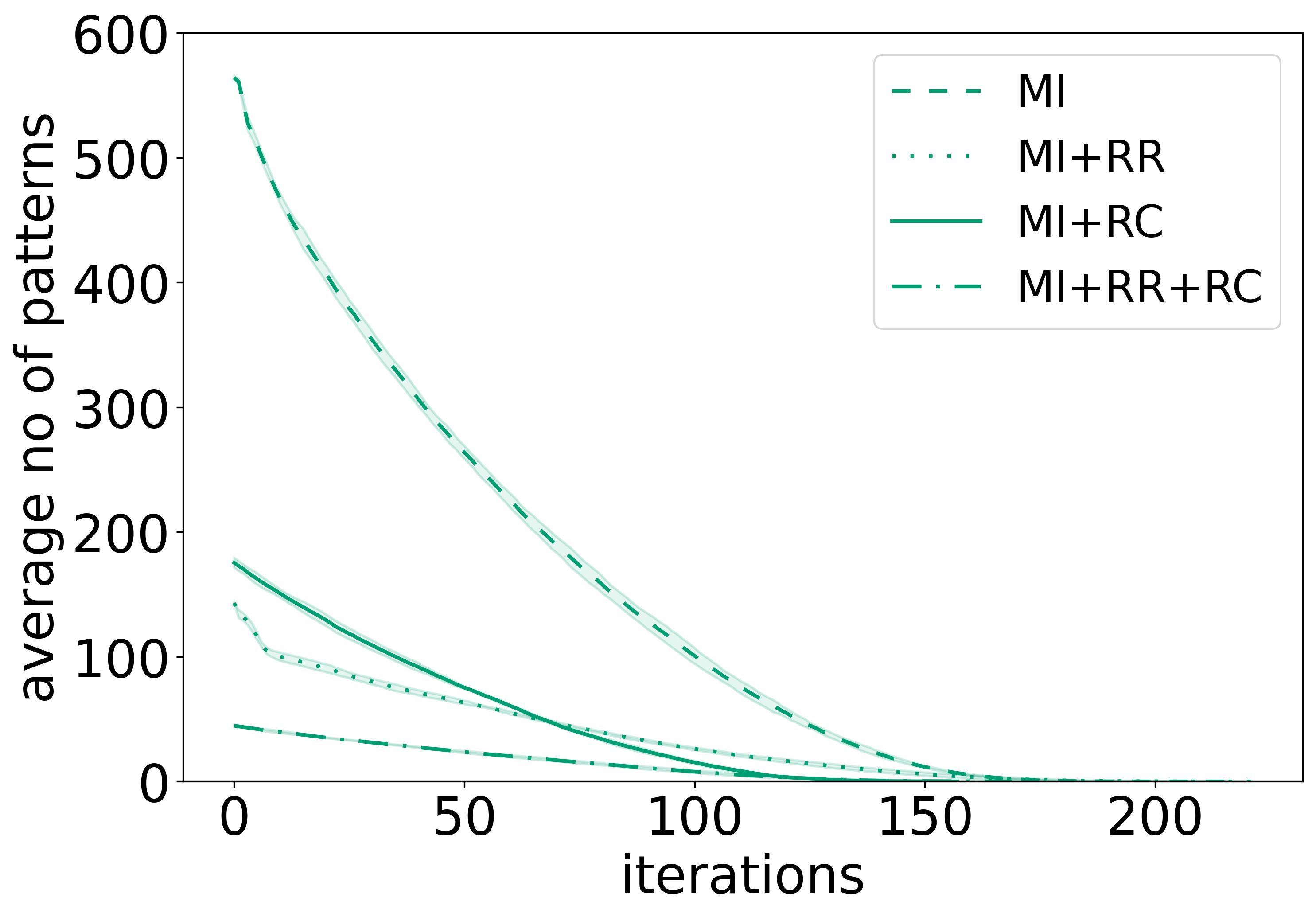}
         \caption{Multiple Similar Input}
         \label{fig:no_patterns_multi_input}
     \end{subfigure}
     \begin{subfigure}[b]{0.32\textwidth}
         \centering
         \includegraphics[width=\linewidth]{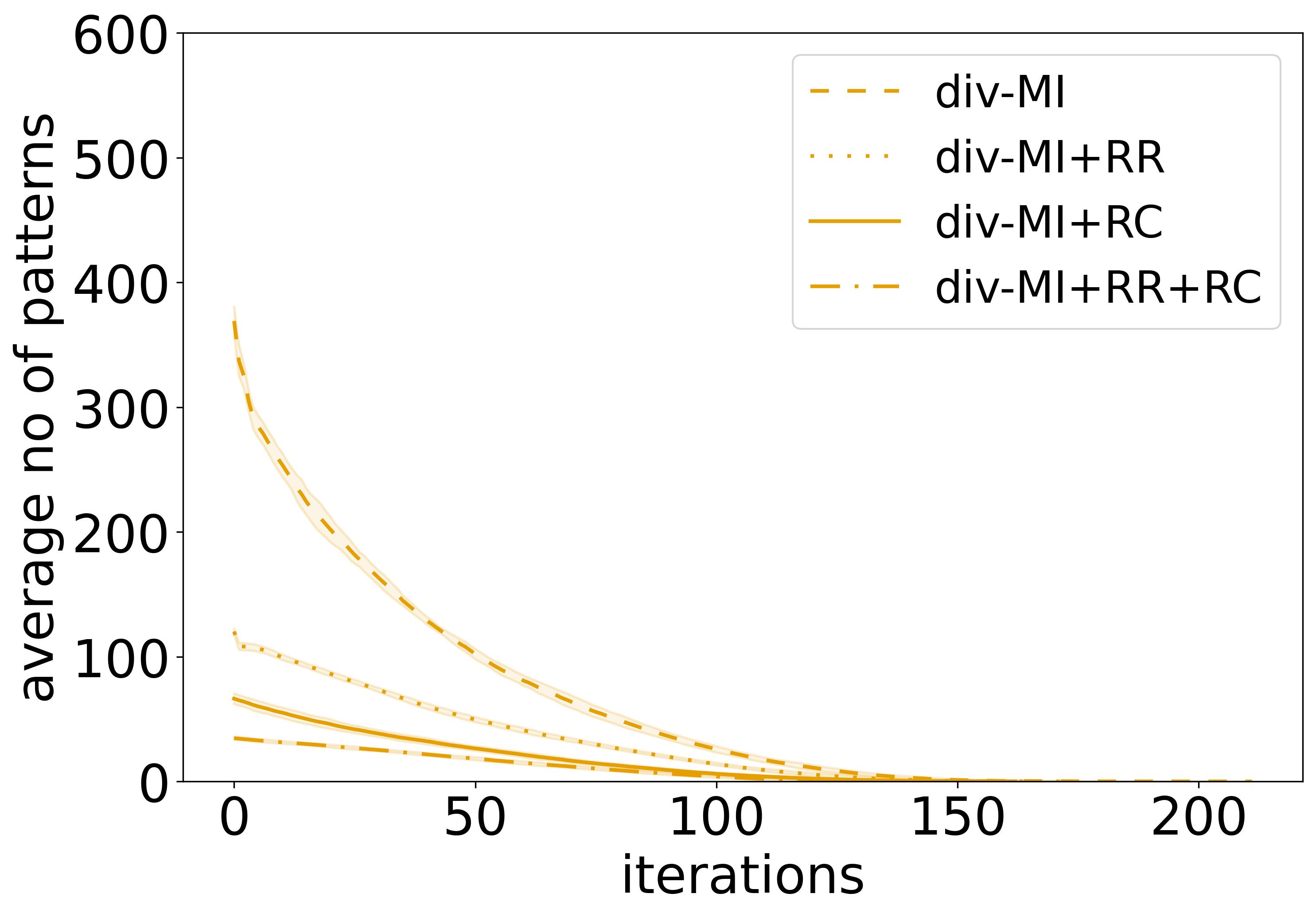}
         \caption{Multiple Diverse Input}
         \label{fig:no_patterns_multi_div_input}
     \end{subfigure}
     \caption{Average number of patterns available for the whole map during training across experimental settings.}
     \label{fig:no_of_patterns_comp}
 \end{figure*} 

We used our simplified automated playing agent to measure the playability of the generated levels. Figure~\ref{fig:playable} compares the percentage of playable levels from different experiments. The playability comparison shows that single- and multiple-input experiments overall generated a higher number of playable levels compared to the highly diverse multiple-input experiments. The lowest-performing experimental setup (div-MI+RR+RC) employs multiple and diverse inputs, removes the rare patterns, and is trained from a random starting state. This finding is expected, as this experiment is the most constrained during training and the hardest to solve. 
The graph clearly shows that div-MI+RR+RC has the highest number of contradictions instead of generating unplayable levels, which showcases that it did not learn to avoid contradictions easily. 

To understand the effect of removing rare patterns, table~\ref{tab:mean_adjaceny} shows the mean values of the number of adjacent patterns for each cell (a proxy for how restrictive/constrained the space is after each selection). We also show the total number of available patterns to the RL agent (the size of the action space) in table~\ref{tab:pattern_numbers}. We excluded patterns that have zero adjacent patterns in any direction, as these patterns lead to contradictions, so the RL usually learns to never use them. Looking at table~\ref{tab:mean_adjaceny}, we notice that the diverse multiple inputs with removing rare patterns has the lowest average compared to all the other configurations. This means that when a tile gets selected, it restricts all the following cells to a small set of patterns, although it has the biggest action space compared to all the others when we remove rare patterns (table~\ref{tab:pattern_numbers}). Another noticeable thing is that the multiple input with removing rare patterns has the highest average number of adjacent patterns, which explains why it leads to higher playability, as it is less constrained after selecting each pattern.


We compare the levels generated using basic WFC and PCGRL alone. While the basic WFC generated visually aesthetic levels but failed to optimize playability. Using multiple inputs managed to generate a small number of playable levels (less than 2\%). But using single and diverse inputs, it failed badly in terms of playability (0\%). Levels generated using PCGRL failed in terms of the visual aspect, making the levels look messy (Figure~\ref{fig:pcgrl_levels}), but on the other hand, they had a relatively higher playability score (15\%). WFCRL also beats the standard PCGRL in terms of inference time. In WFCRL, placing one pattern leads to collapsing multiple cells at each step and terminates once all cells are collapsed, while standard PCGRL modifies one tile at each step and keeps iterating until a goal state or iteration limit is reached.  

\begin{table}
    \centering
    \begin{subfigure}[t]{0.47\linewidth}
        \centering
        \begin{tabular}{|l|c|c|}
            \cline{2-3}
            \multicolumn{1}{l|}{} & No RR & RR \\
            \hline
            SI & 4.58  & 4.59  \\
            MI & 4.80 & 5.03  \\
            div-MI & 4.06 & 3.67 \\
            \hline
        \end{tabular}
        \caption{\#Adjacent patterns}
        \label{tab:mean_adjaceny}
    \end{subfigure}
    \hfill
    \begin{subfigure}[t]{0.47\linewidth}
        \centering
        \begin{tabular}{|l|c|c|}
            \cline{2-3}
            \multicolumn{1}{l|}{} & No RR & RR \\
            \hline
            SI & 306 & 100 \\
            MI & 645 & 177 \\
            div-MI & 614 & 196 \\
            \hline
        \end{tabular}
        \caption{\#Patterns}
        \label{tab:pattern_numbers}
    \end{subfigure}
    \caption{The average number of adjacent patterns and the total number of available patterns for all the experiments.}
    \label{tab:adjaceny_pattern_values}
\end{table}

Figure~\ref{fig:no_of_patterns_comp} shows the number of available patterns at every step during training. It is very noticeable that the number of available patterns decreases by a large amount when either random collapse or removing rare patterns is used. This explains why it leads to low playability in general, as the agent during training has fewer choices to pick from at every step, which might not always lead to playability. Of course, using diverse input also becomes worse since the remaining choices might be from different input levels that do not have any connections, which leads to a contradiction.

\begin{figure}
    \centering
    \includegraphics[width=0.95\linewidth]{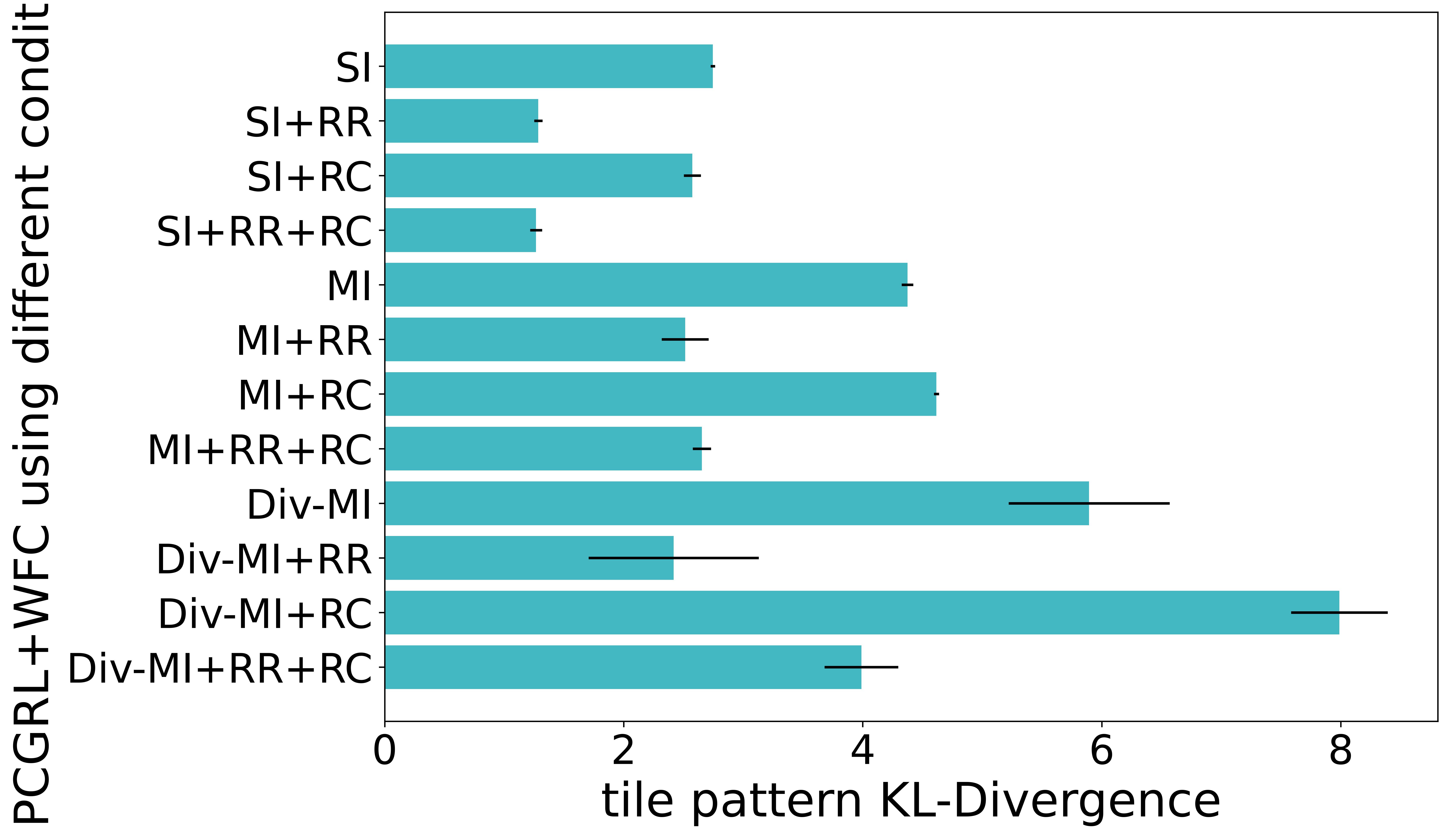}
    \caption{Diversity of playable levels generated using different conditions: single input (SI), multiple inputs (MI), highly diverse multiple input (Div-MI), excluding rare patterns (RR), random collapsed starting state for training (RC).}
    \label{fig:diversity}
\end{figure}

To understand more about the generated levels and how diverse they are from each other, we compare the diversity of the generated levels. We use TP-KLDiv as a diversity metric on the playable levels using a $3 \times 3$ window. Figure~\ref{fig:diversity} displays the diversity of the playable levels across different experiments. An overall noticeable trend is that levels generated using multiple input frames have higher diversity values compared to the single-input experiments. This is the effect of the larger action space. Multiple inputs give more options to choose from, which helps the algorithm to generate levels different from each other, as seen from the examples in Figures~\ref{fig:single_input}, \ref{fig:multiple_input}, and \ref{fig:div_multiple_input}. An interesting observation is that for each of the single, multiple, and diverse multiple input experiments, diversity is higher when rare patterns are included, and the diversity decreases when rare patterns are excluded from the dataset. Removing the rare patterns leads to a decrease in the size of the action space, as shown in Figure~\ref{fig:no_of_patterns_comp}. Therefore, the agent is left with a smaller number of options and tends to repeat similar actions. This trend is also visible in the generated levels.

\section{User Survey}

We conducted a survey to understand users' preferences on the levels generated using different methods. We used levels generated using basic WFC, PCGRL, and our two models from the WFCRL framework: MI+RR (highest playability) and Div-MI+RC (highest diversity). As we are interested in the visual preference of the user, we used images of levels and the user is not required to play the levels (playing the levels takes longer and can make the user less interested). The survey consists of 5 rounds; in each round, the user is given images of two randomly selected levels and asked a few preference questions where they need to select either one of the levels (left or right), both, or none. The questions are the following:
\begin{itemize}
    \item Which level do you find more visually aesthetic?
    \item Which level looks more playable to you?
    \item Which level you think has repetitive layout patterns (structures, elements placement, etc)?
    \item Which one has higher structural similarity with Lode Runner levels?
\end{itemize}
To make sure the users are familiar with the game and its structure, we give an introduction to the game by showing a gameplay video, and we also show multiple example levels to make the user familiar with the structural aesthetic look of Lode Runner levels.

\begin{table}
\centering
\resizebox{\linewidth}{!}{
\begin{tabular}{|l|r | l|r|}
 \hline
    \multicolumn{2}{|c|}{Gaming frequency } & \multicolumn{2}{c|}{Familiarity with Lode Runner}\\
 \hline
    Everyday     & 38.88\%    & Played many times & 5.55\% \\
    Few times a week     & 38.88\%    & Played few times & 38.88\% \\
    Once a week     & 11.11\%    & Familiar but never played &  27.77\%  \\
    Rarely     & 11.11\%    & Not familiar &  27.77\%  \\
     \hline
\end{tabular}
}
\caption{Demographic details of users.}
\label{tab:demographics}
\end{table}

We ran the online survey for a week, and we got 18 users. We collected some demographics to understand more about our users. We noticed that a major part of the users are familiar with Lode Runner and play games frequently (as shown in table~\ref{tab:demographics}). This was expected since the researcher network and groups they posted in are more directed towards video game players.

\begin{figure}
    \centering
    \includegraphics[width=0.95\linewidth]{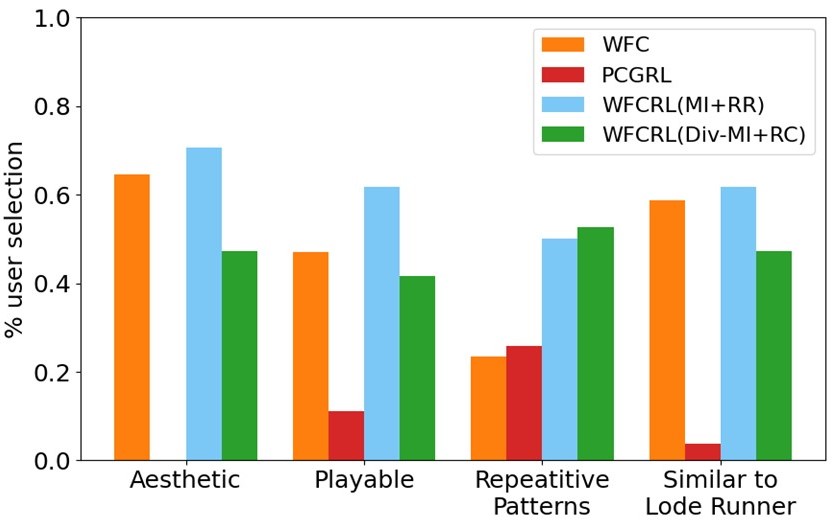}
    \caption{User study results from 18 users comparing two of our trained models to the WFC algorithm and PCGRL algorithm alone. Our two models are: MI+RR, which has the highest playability, and Div-MI+RC, which has the highest diversity.}
    \label{fig:res_survey}
\end{figure}

Figure~\ref{fig:res_survey} shows the results from the survey; the most noticeable point is that PCGRL was not superior in any of the questions except for the repetitive patterns. We believe that is due to the noisy structure of the generated levels. The surprising thing is that users did not think that the levels from PCGRL are more playable than WFC (which, in reality, it is). We believe that due to the noise-like levels of the PCGRL levels (figure~\ref{fig:pcgrl_levels}), the users might get overwhelmed and think that it is not playable even when it is. An interesting aspect is Div-MI+RC model was not considered similar to Lode Runner levels, although looking at the levels (figure~\ref{fig:div_mi_rc}), they are similar to the training data. We believe that the users might not consider dense-looking levels where a lot of solid tiles are everywhere as a typical Lode Runner level, and consider other levels with platforms and ladders more typical. 

Only one of our models (MI+RR) managed to beat the normal WFC in playability. That is interesting since all the generated levels by WFC are unplayable, while most of the generated levels by our model are playable, so we expected that both our model and PCGRL would have higher playability. This might be due to how the question was framed, as to which level is more playable. Some users might have flagged nearly playable WFC levels as playable, causing the values to be close to each other. Another reason can be the puzzle aspect of Lode Runner levels; some users might consider a difficult level as unplayable, as they could not solve the level.

\section{Limitations}
One drawback of our method is the simplified playability measure. To make the measure faster, we exclude the digging ability of the agent, enemy movement, and do not consider the order of gold collection. Without digging, the RL agent will learn to avoid selecting such patterns that require digging to reach gold. Also, excluding the digging ability makes some levels marked as unplayable, which can be playable if digging is applied. Our playability does not simulate the enemy movements; therefore, the presence or high number of enemies does not affect the gameplay. In the original game, some levels strictly required following a specific order of collecting gold. As we do not consider the order of reaching the gold, the deadlock situations are not evaluated here.

We tested our method on Lode Runner only. Therefore, we do not have evidence about its applicability to other domains. Our method relies on spatial dependencies of the tiles; in Lode Runner, the tiles are highly dependent on each other, like a ladder connected to a rope or a brick, gold placed on a brick, etc. We are unsure about the effectiveness of our method where the tiles have less or no interdependencies. We are also unsure about whether the method is applicable to 3d game domains or not; these require further investigation.

\section{Conclusion}

In this work, we explored combining WFC with PCGRL in order to gain the advantages of each method. The output framework was able to combine both of their features compared to using only one of the methods. In our proposed method, WFC managed to constrain the space for the PPO-based PCGRL agent to ensure the generated levels have a similar pattern distribution compared to levels generated by PCGRL only (see Figure~\ref{fig:pcgrl_levels}). While WFC alone generated visually pleasing levels but failed to optimize playability. On the other hand, PCGRL managed to increase the playability of the generated content compared to the basic WFC but did not optimize the visual aspect. We also explored the effect of three different hyperparameters on the framework: input data, learned patterns, and starting state. 
 
For input data, we found out that using more than one input level helps to increase the playability overall, as long as these levels are similar to each other in structure. This allows the algorithm to move between the different patterns easily, as they are similar to each other. On the other hand, having diverse inputs increases the diversity of the generated levels but leads to less playable levels, as the framework faces challenges in finding connections between different types of patterns. We argue that with a smoother gradient between diverse levels--such as levels that combine both styles--our method would yield even better results.

For the learned patterns, removing rare patterns causes a decrease in the diversity of the generated levels due to the smaller number of available patterns. But it helps increase the number of playable levels. We believe that removing rare patterns helped the framework to focus on the most common patterns that usually lead to playable levels, rather than the unique patterns that appear rarely in the input levels.

Finally, for the starting state, starting from a partially collapsed state did not show much difference in playability, and little improvement in diversity. The trained models are more robust to the starting state and can actually find playable levels more easily when not starting from an empty state. We believe random collapse is a key feature to have more generic policies that can work across different games and will help in transfer learning.

We also ran a user study with $18$ users to understand how our method is perceived by users. We found out that although we have more playable levels than WFC levels, users did not give our method a higher score on the playability scale in comparison. We believe the puzzle nature of Lode Runner might have influenced the users not to notice which level is more playable. This might be an interesting avenue to tackle in the future on how to generate levels that are easier to perceive as playable.

\bibliography{aaai24}
\end{document}